\documentclass{article}
\usepackage{spconf,amsmath,graphicx}
\usepackage{subfigure}
\usepackage{CJKutf8}
\usepackage{enumitem}

\def\L{{\cal L}}

\title{A Novel Chinese Dialect TTS Frontend with Non-Autoregressive Neural Machine Translation}
\name{Junhui Zhang, Wudi Bao, Junjie Pan, Xiang Yin, Zejun Ma}
\address{Bytedance AI-Lab}

\begin{document}

\maketitle
\begin{abstract}
 
Chinese dialects are different variations of Chinese and can be considered as different languages in the same language family with Mandarin. Though they all use Chinese characters, the pronunciations, grammar and idioms can vary significantly, and even local speakers may find it hard to input correct written forms of dialect. Besides, using Mandarin text as text-to-speech inputs would generate speech with poor naturalness. In this paper, we propose a novel Chinese dialect TTS frontend with a translation module, which converts Mandarin text into dialectic expressions to improve the intelligibility and naturalness of synthesized speech. A non-autoregressive neural machine translation model with various tricks is proposed for the translation task. It is the first known work to incorporate translation with TTS frontend. Experiments on Cantonese show the proposed model improves 2.56 BLEU and TTS improves 0.27 MOS with Mandarin inputs.

\end{abstract}

\section{Introduction}

In Chinese text-to-speech (TTS) system, frontend serves to convert text into linguistic features as inputs of the acoustic model, and greatly affects the pronunciation accuracy and prosody naturalness of the synthesized speech. Typical Chinese TTS frontend consists of a series of text processing components with neural network techniques, including text normalization \cite{sproat2016rnn,zhang2020hybrid}, Chinese word segmentation with part-of-speech tagging \cite{huang2015bidirectional,dong2016character,devlin2018bert,radford2018improving}, prosody prediction \cite{chu2001locating,pan2019mandarin,lu2019self,clark2020electra}, and grapheme-to-phoneme (G2P) conversion \cite{yolchuyeva2020transformer}.

Dialects are local varieties of Chinese widely used in different regions of China. Although dialects and Mandarin share the same written symbols, i.e. Chinese characters, their written forms may differ significantly. Given a Mandarin text for dialect speech synthesis, not only the pronunciation of each character may be totally different from Mandarin, but the word usage and grammar may not be idiomatic as well. For example, in character level, ``\begin{CJK}{UTF8}{bkai}他\end{CJK}" means ``him" in Mandarin, but Cantonese uses ``\begin{CJK}{UTF8}{bkai}佢\end{CJK}" instead; Shanghai dialect uses ``\begin{CJK}{UTF8}{gkai}侬\end{CJK}" representing ``you", which is ``\begin{CJK}{UTF8}{bkai}你\end{CJK}" in Mandarin. ``\begin{CJK}{UTF8}{bkai}他\end{CJK}" and ``\begin{CJK}{UTF8}{bkai}你\end{CJK}" in Cantonese and Shanghai dialect have different meanings or usage. In word and grammar level, such difference becomes even bigger. As a result, reading a Mandarin text directly with dialect pronunciations would produce unnatural and unidiomatic speech. 

Typical dialect TTS frontend has the same pipeline as Mandarin with each component designed or trained for a specific dialect. Taking G2P as an example, each dialect has its own dictionaries and polyphone prediction model. For the acoustic model in dialect TTS, the training data are recorded with dialect scripts, and the model can generate natural dialect speech with dialect texts. However, most users, including native speakers, can only write Mandarin text instead of dialect because the written form of dialects is usually the knowledge for linguistic experts. We find, from our dialect TTS service, that most users input Mandarin text inputs for dialect TTS. In this way, the acoustic model of dialect TTS suffers from a mismatch of text inputs between training and inference, leading to a degradation of naturalness of the synthesized dialect speech. Zhang \cite{2020arXiv201011489Z} proposes to use Shanghai dialect speech to tune the acoustic model and vocoder respectively. The results show that although the pronunciations of the synthesized speech are satisfying, the intelligibility is deeply undermined.

To lower the usage barrier for native and non-native users to generate natural dialect speech, we propose a dialect TTS frontend including a text translation module to convert the input Mandarin text to native dialect text for natural dialect speech generation. Various neural machine translation (NMT) models can be adopted for this task. Autoregressive transformer (AT) \cite{vaswani2017attention} can gain extremely high translation performance but with massive computation. Wu \cite{wu2020lite} proposes a lightweight AT using multi-head self-attention and depth-wise separable convolution attention. In addition, non-autoregressive transformer (NAT) \cite{gu2017non} has also received wide attention. Due to its parallelization property, NAT has faster inference speed but with a relatively lower performance. Qian \cite{qian2020glancing} proposes GLAT with glancing sampling and incremental learning strategies and greatly improves the translation performance compared with other NAT models.

In this paper, we design our translation model based on GLAT and propose some tricks to further improve its performance. In Section 2, we describe our dialect TTS frontend pipeline, translation module, and optimization methods. In Section 3, we conduct experiments on Mandarin-Cantonese text to prove the effectiveness of our proposed model. In Section 4, we summarize the progress made in this paper and discuss research priorities for the future works. The contributions of this paper are summarized as follows:
\begin{itemize}[itemsep=2pt,topsep=0pt,parsep=0pt]
\item This is the first known work to incorporate translation into Chinese dialect TTS frontend.
\item We introduce various tricks to optimize our translation model to gain better performance.
\end{itemize}

\section{Methods}

\subsection{Chinese Dialect TTS Frontend Pipeline}
We add a translation module in a typical dialect TTS frontend, and the proposed pipeline is shown in Figure 1. The input Mandarin text is first sent to the translation module. In order to preserve the untranslated content, we adopt a special-token replacement technique which will be detailly explained in Section 2.2.6. After translation, the processed text goes through the remaining frontend components before converted into linguistic information as the frontend result.

\begin{figure}
  \centering
  \includegraphics[width=6cm]{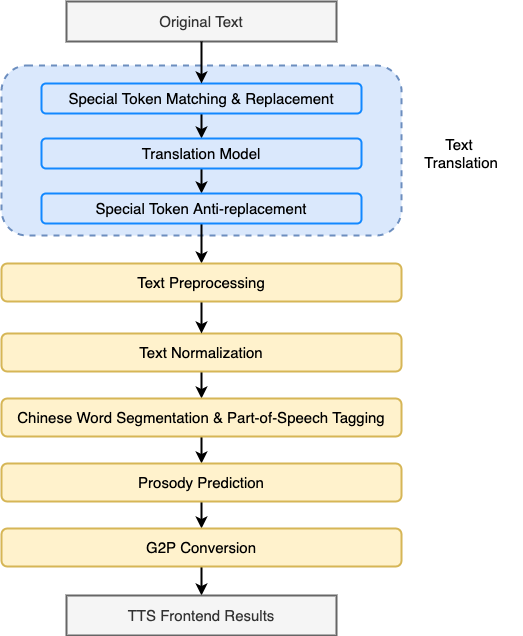}
  \caption{Proposed dialect TTS frontend pipeline.}
  \label{fig1:TTS frontend pipeline}
\end{figure}

\subsection{Translation Module}

Since the translation module needs to be integrated into the TTS frontend, it must have fast inference speed and be small enough to be deployable. The commonly used NMT models based on AT transformers cannot reach such requirements. Hence, we develop our model based on a lightweight NAT model GLAT \cite{qian2020glancing}. As shown in Figure 2, the entire model is composed of the following 4 modules: a multi-branch transformer encoder, a multi-branch NAT decoder, a length predictor and a translation predictor. The model details and proposed optimization tricks are mentioned in the sub-sections.

\begin{figure}
  \centering
  \includegraphics[width=\linewidth]{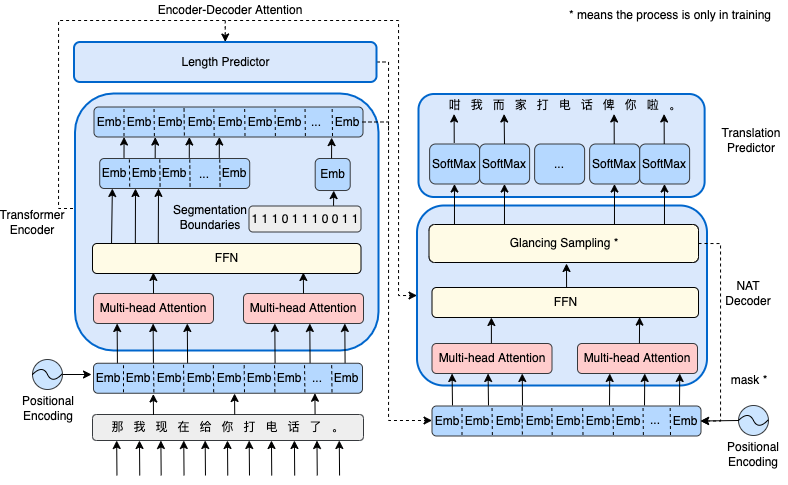}
  \caption{Translation model. Encoder outputs and word segmentation embeddings are concated together. Output sentence lengths are predicted by encoder-decoder attention. Decoding is required twice during training but once in inference.}
  \label{fig2:translation module}
\end{figure}

\subsubsection{GLAT}

GLAT \cite{qian2020glancing} is a transformer-based NAT translation model with optimized training strategies. Typical NAT models can only learn limited word dependencies since they output predictions completely in parallel. To solve the problem, GLAT adopts a two-pass decoding strategy during training. The first pass is to get an output prediction directly. Then, a proportion of target words will be sampled to ``tell" the decoder by replacing the decoder input. The second pass is to predict the remaining unsampled words. During inference, only the first pass decoding is performed, which ensures the inference speed is as fast as a typical NAT model.

\subsubsection{Multibranch}

In order to reduce the computation to improve training efficiency, we adopt the multi-branch mechanism \cite{wu2020lite} where the original self-attention is replaced by multi-branch convolution attention and self-attention. In our model, we simply use multi-head self-attention since it is powerful enough to capture both global and local features. Specifically, if the number of branches is \textit{n}, the input embedding is divided into \textit{n} parts. This strategy is adopted in the multi-head self-attention layers in both encoder and decoder. The attented outputs in each branch are concated together following a FFN layer.

\subsubsection{Word Segmentation Embedding}

Our model uses characters instead of words as token inputs for better performance and a smaller dictionary size thus a smaller model size. However, word segmentation information is still beneficial for the understanding of sentence contents since one character may have multiple meanings in Chinese dialects. Hence, in order to help the model learn the information and dependencies on word level, we add extra word segmentation boundary information into the decoder. We simply use 0 and 1 to indicate the boundaries.

\subsubsection{Data Augmentation}

We find that GLAT suffers from predicting duplicated characters and is not robust enough on sentences that are too short or too long. Auto-regressive models, on the other hand, predict outputs based on preorder words, and can effectively avoid this problem. As a result, we first train a high-performance auto-regressive model, collect Mandarin text corpus with various lengths and contents, and use the model to translate Mandarin text to dialect text as the augmented dataset. This data augmentation approach would greatly improve the robustness of the NAT model.

\subsubsection{Alignment Loss}

A visualization of attention weights can show the soft alignment of source words and target words \cite{bahdanau2014neural}. However, our model only has a character-level alignment since we use characters as token. To get a more reasonable alignment, we introduce alignment loss to help the model further understand the sentence meaning at word level. We train GIZA++v2 \cite{brown1993mathematics} \cite{och2003systematic} with the full amount of parallel text corpus to generate target alignments as training objective. The alignment loss is the following
\begin{equation}
  L = \ell(x, y) = \{l_1, ..., l_N\}^T, l_n = (x_n-y_n)^2
  \label{eq1}
\end{equation}
where \textit{N} is batch size; \textit{x}, \textit{y} are predicted and target alignment.

\subsubsection{Special Token Replacement}
Considering the diversity of user input texts, we need to ensure that some untranslated contents are output as original, such as symbolic expressions, common abbreviations, web addresses, etc. Since they may not appear in the training set, they are typically recognized as unknown tokens and then predicted as irrelevant text in traditional translation models. We propose to match these contents by regular expression in the training corpus, and replace with a special token $\langle rep\rangle$. In this way, the model can learn the correspondence between the special tokens in source and target sentences. During inference, we replace the predicted special tokens with the original strings so that the original untranslated contents can be preserved. For example, a source sentence ``\begin{CJK}{UTF8}{gkai}这个男生是\end{CJK}Sam" (This boy is Sam) becomes ``\begin{CJK}{UTF8}{gkai}这个男生是\end{CJK}$\langle rep\rangle$" and is translated to ``\begin{CJK}{UTF8}{gkai}呢个男仔系\end{CJK}$\langle rep\rangle$", which is then reorganized as the final output ``\begin{CJK}{UTF8}{gkai}呢个男仔系\end{CJK}Sam".

\section{Experiments and Results}
We mainly conduct experiments on the translation model to prove the effectiveness of the proposed tricks. The overall dialect TTS system with the original and proposed frontend is further evaluated to show the improvement brought by the proposed method. The translation model with different tricks are objectively evaluated in Section 3.3.1 and subjectively evaluated in Section 3.3.2. The entire TTS system is evaluated subjectively in Section 3.3.2.
\subsection{Dataset}
Cantonese is a distant dialect with Mandarin on word usage and grammar, and is representative in dialect TTS problems. In our experiments, Cantonese is used as the target dialect for our proposed dialect TTS frontend. The training dataset for the translation model includes 210k Mandarin-Cantonese parallel corpus. Data augmentation in Section 2.2.4 generates 90k additional translated text pairs by an auto-regressive translation model trained on the 210k corpus.
\subsection{Experiment Settings}
Our model has one block of encoder and decoder and the input/output dimension $d_{model}$ is 300. Word2vec pre-trained embeddings\footnote{pre-trained from https://github.com/Embedding/Chinese-Word-Vectors.} are used to initialize the character input embeddings in training. The source and target embedding dictionaries are combined together so that same token would have same embedding in Mandarin and Cantonese. Other settings of the translation model remain the same as original GLAT.
 
 For the subjective evaluation of the TTS system, we use Cantonese TTS with and without the translation module for speech synthesis. Tacotron \cite{wang2017tacotron} and WaveRNN \cite{kalchbrenner2018efficient} are used as as the acoustic model and vocoder, both of which are trained on our internal Cantonese speech dataset.

\subsection{Results and Analysis}
\subsubsection{Objective Evaluation}
We first compare the performance of the translation model with each optimization trick proposed in Section 2.2.2 to 2.2.5 using BLEU \cite{papineni2002bleu}. The models include original GLAT model (baseline), GLAT with each trick and proposed GLAT model with all tricks. Besides, an AT translation model is trained as another baseline. Real-time factor (RTF) is used for inference speed calculation. The results are shown in Table 1.

\begin{table}
  \caption{Models with different optimization methods. Each model is GLAT baseline with one trick, and the proposed model is GLAT baseline with all tricks above.}
  \label{tab:model bleu}
  \centering
  \begin{tabular}{l|l|c}
    \hline
    \multicolumn{1}{c|}{\textbf{Model}} & \multicolumn{1}{c|}{\textbf{BLEU}} & \multicolumn{1}{c}{\textbf{RTF}} \\
    \hline
    GLAT (baseline)                   & 63.10~~~         & - \\
    \hline
    word seg embedding                & 63.54(+0.44)~~~      & - \\
    data augmentation                 & 64.89(+1.79)~~~       & - \\
    multi-branch                       & 63.46(+0.36)~~~       & - \\
    alignment loss                    & 63.77(+0.67)~~~       & - \\
    \textbf{proposed model}           & \textbf{65.66(+2.56)}~~~   & \textbf{0.002} \\
    \hline
    AT \cite{vaswani2017attention} (baseline)        & 67.02~~~  & 0.038 \\
    \hline
  \end{tabular}
  
\end{table}

Overall, the proposed model improves 2.56 BLEU compared to the baseline GLAT model. By introducing data augmentation, the model improves 1.79 BLEU and we find it more robust on inputs with different lengths and no longer suffers from producing duplicated characters. Training the model with a supervised alignment loss brings an improvement of 0.67 BLEU, and the attention weight matrices with and without the alignment loss are shown in Figure 4. The original attention weights have ambiguous and incorrect character-level correspondence, but the model with alignment loss produces attention weights with meaningful word-level correspondence. Multi-branch and word segmentation embedding help improve 0.36 and 0.44 BLEU respectively and reduce the model size by approximately 7\%. Besides, compared with the AT translation model, the proposed model has a close BLEU score performance but an approximately 20 times faster inference speed.

\begin{figure}
\centering
\subfigure[Attention without alignment loss]{\label{fig:a}\includegraphics[width=4.2cm]{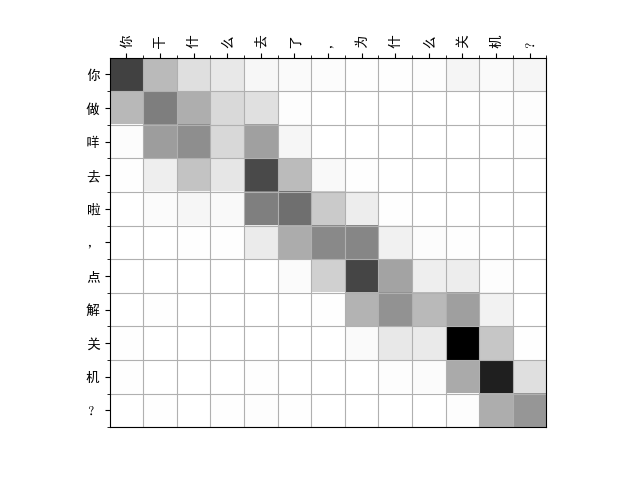}}
\subfigure[Attention with alignment loss]{\label{fig:b}\includegraphics[width=4.2cm]{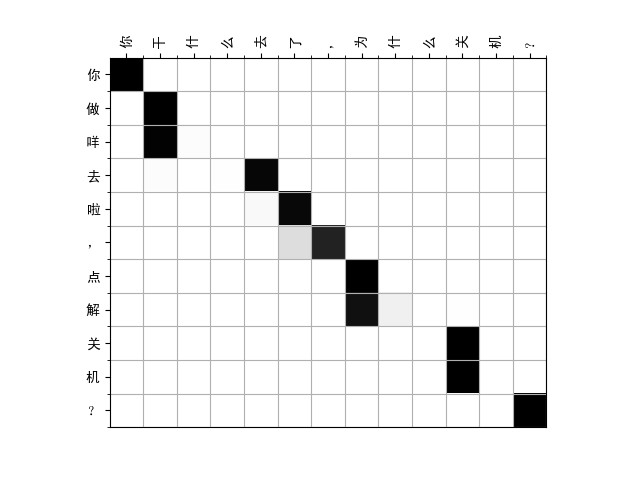}}
\caption{Visualization of attention matrices with and without supervised alignment training.}
\end{figure}

\subsubsection{Subjective Evaluation}

We carefully design a set of subjective evaluation criterion called Translation Mean Opinion Score (TMOS) described in Table 2. Although BLEU is the most widely used method in machine translation evaluation, it still has some disadvantages. First, since BLEU is based on n-gram matching, it is easy to ignore grammatical issues. Second, it is determined by whether the words in the candidate appear in the target and would score wrongly on candidates only containing a few key words. Besides, BLEU can only consider one or a few fixed references within the dataset. In contrast, our proposed TMOS can score semantically, which is more in line with subjective perception. In addition, TMOS is conducted by native speakers without fixed reference, so it can take into account different synonyms and expressions.

In TMOS test, 40 Cantonese native speakers are asked to score the translation results from three systems: ground-truth, baseline GLAT, and proposed model. Each participant is required to evaluate 200 predictions for each system, and the result distributions are in Figure 5. The result shows that the translated sentences of proposed model with scores of 4.5 and 5 both improve 3\% compared to original GLAT.

\begin{table}
  \caption{Subjective evaluation criterion.}
  \label{tab:word_styles}
  \centering
  \begin{tabular}{l|c}
    \hline
    \multicolumn{1}{c|}{\textbf{Translation Result}} & \multicolumn{1}{c}{\textbf{Score}} \\
    \hline
    No output/More than half of the output is garbled     & 1.0             \\
    Output semantically irrelevant text                   & 2.0             \\
    Output semantically imperfectly relevant text         & 3.0             \\
    Error-prone direct translation                        & 4.0             \\
    Error-free direct translation                         & 5.0             \\
    \hline
  \end{tabular}
\end{table}

\begin{figure}
  \centering
  \includegraphics[width=\linewidth]{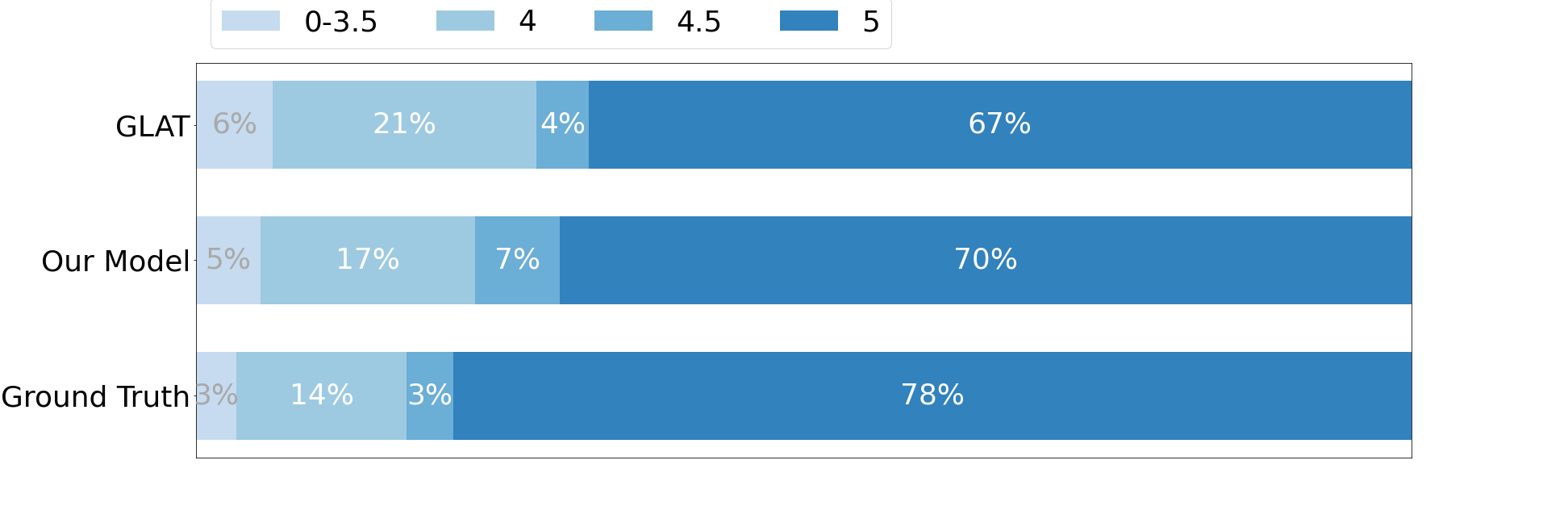}
  \caption{Distribution bar plots of subjective TMOS evaluation.}
  \label{fig5:TMOS}
\end{figure}

\begin{table}[ht!]
  \caption{MOS of synthesized Cantonese speech.}
  \label{tab:word_styles}
  \centering
  \setlength{\tabcolsep}{7mm}{
  \begin{tabular}{c|c|c}
    \hline
     & \textbf{baseline}  & \textbf{proposed}         \\
    \hline
    MOS    &      4.00 ($\pm$0.63)    &    4.27 ($\pm$0.63)      \\
    \hline
  \end{tabular}}
\end{table}

For TTS subjective evaluation, baseline dialect frontend without translation module and proposed frontend are used in the Cantonese TTS systems to generate speech samples. The baseline directly reads Mandarin text with Cantonese pronunciations. 20 Cantonese native speakers are asked to evaluate the naturalness and intelligibility of the synthesized speech using Mean Opinion Score (MOS). The evaluation results in Table 3 show that the synthetic quality of the proposed dialect frontend is significantly better than baseline, and the synthesized speech has an obvious improvement in prosody and naturalness. Audio samples are available online.\footnote{demo page: https://zhangjh915.github.io/icassp-2023-dfe/}

\section{Conclusions}

We propose a novel Chinese dialect TTS frontend with a non-autoregressive text translation module and evaluate it on the Mandarin-Cantonese dataset. Our proposed model improves 2.56 BLEU on the translation task. Dialect TTS with the proposed frontend can generate speech with better comprehensibility and naturalness, indicated by a 0.27 improvement in MOS. For future work, we will further improve the dialect frontend by developing a detection mechanism to decide whether the input needs to translate to dialect text.

\vfill\pagebreak

% -------------------------------------------------------------------------
\bibliographystyle{IEEEbib}
% \bibliography{refs}

\end{document}